\documentclass[letterpaper, 10 pt, conference]{IEEEtran}

\IEEEoverridecommandlockouts
\usepackage{geometry}
\usepackage{cite}
\usepackage{amsmath,amssymb,amsfonts}
\usepackage{amsmath}
\usepackage{graphicx}
\usepackage{textcomp}
\usepackage{multirow}
\usepackage{xcolor}
\usepackage{algorithm}  
\usepackage{booktabs}
\usepackage{algorithm,algpseudocode}
\usepackage{soul,color}

\algnewcommand{\algorithmicforeach}{\textbf{for each}}
\algdef{SE}[FOR]{ForEach}{EndForEach}[1]
{\algorithmicforeach\ #1\ \algorithmicdo}
{\algorithmicend\ \algorithmicforeach}
\usepackage{amsmath}
\usepackage{threeparttable}
\usepackage{subfigure}
\usepackage{url}

\usepackage{amsmath}
\usepackage{amsfonts}
\usepackage{amssymb}
\usepackage{algorithm,algpseudocode}
\usepackage{hyperref}
\makeatletter
\newcommand{\algmargin}{\the\ALG@thistlm}
\makeatother
\newlength{\whilewidth}
\algdef{SE}[parWHILE]{parWhile}{EndparWhile}[1]
{\parbox[t]{\dimexpr\linewidth-\algmargin}{%
		\hangindent\whilewidth\strut\algorithmicwhile\ #1\ \algorithmicdo\strut}}{\algorithmicend\ \algorithmicwhile}%
\algnewcommand{\parState}[1]{\State%
	\parbox[t]{\dimexpr\linewidth-\algmargin}{\strut #1\strut}}

\def\BibTeX{{\rm B\kern-.05em{\sc i\kern-.025em b}\kern-.08em
    T\kern-.1667em\lower.7ex\hbox{E}\kern-.125emX}}

\title{\LARGE \bf
	ReVoLT: Relational Reasoning and Voronoi Local Graph Planning\\ for Target-driven Navigation\\
} 

\author{\IEEEauthorblockN{Junjia Liu$^1$$^3$, Jianfei Guo$^2$$^3$, Zehui Meng$^3$, Jingtao Xue$^3$}
	\IEEEauthorblockA{$1$ \textit{Department of Mechanical and Automation Engineering}, \textit{The Chinese University of Hong Kong}\\
		$2$ \textit{School of Automation Science and Engineering}, \textit{Xi'an Jiaotong University}\\
		$3$ \textit{Application Innovate Laboratory (2012 Laboratories)}, \textit{Huawei Technologies Co., Ltd.}\\
		Beijing, 100038, China \\
		jjliu@mae.cuhk.edu.hk, ventus@stu.xjtu.edu.cn, \{mengzehui, xuejingtao\}@huawei.com
	}
}

\begin{document}
\maketitle
\thispagestyle{empty}
\pagestyle{empty}

\begin{abstract}      
	Embodied AI is an inevitable trend that emphasizes the interaction between intelligent entities and the real world, with broad applications in Robotics, especially target-driven navigation. This task requires the robot to find an object of a certain category efficiently in an unknown domestic environment. Recent works focus on exploiting layout relationships by graph neural networks (GNNs). However, most of them obtain robot actions directly from observations in an end-to-end manner via an incomplete relation graph, which is not interpretable and reliable. We decouple this task and propose \textit{ReVoLT}, a hierarchical framework: (a) an object detection visual front-end, (b) a high-level reasoner (infers semantic sub-goals), (c) an intermediate-level planner (computes geometrical positions), and (d) a low-level controller (executes actions). \textit{ReVoLT} operates with a multi-layer semantic-spatial topological graph. The reasoner uses multiform structured relations as priors, which are obtained from combinatorial relation extraction networks composed of unsupervised GraphSAGE, GCN, and GraphRNN-based Region Rollout. The reasoner performs with Upper Confidence Bound for Tree (UCT) to infer semantic sub-goals, accounting for trade-offs between \textit{exploitation} (depth-first searching) and \textit{exploration} (regretting). The lightweight intermediate-level planner generates instantaneous spatial sub-goal locations via an online constructed Voronoi local graph. The simulation experiments demonstrate that our framework achieves better performance in the target-driven navigation tasks and generalizes well, which has an 80\% improvement compared to the existing state-of-the-art method. The code and result video will be released at \href{https://ventusff.github.io/ReVoLT-website/}{https://ventusff.github.io/ReVoLT-website/}. 
	
\end{abstract}


\section{Introduction}
Finding objects in complex houses efficiently is a prerequisite for domestic service robots. Robots need to reason and make dynamic decisions along with interacting with the real-world environment. \textit{Embodied AI}, proposed by Matej Hoffman and Rolf Pfiefer\cite{hoffmann2012implications}, suggests that to truly understand how the human brain works, a brain should be embedded into a physical body, and let it explore and interact with the real world. Among all the work practicing \textit{Embodied AI} in recent years, target-driven navigation (TDN) is one of the most feasible and essential tasks, which combines techniques in both machine learning and robotics, and is widely applicable for scenarios such as domestic service robots. It typically requires the robot to find a target object of a certain category in an unknown scene, demanding both high efficiency and success rate. Hence, the key problems of the TDN task are generalizing across unknown domains and exploring efficiently.

Traditional Simultaneous Localization and Mapping (SLAM) pipeline has already handled TDN to some extent\cite{activeneuralslam}, but there are still numerous problems lying in its major modules. First, it remains troublesome for SLAM-based methods to acquire and maintain a lifelong updating semantic map, which demands accurate sensors and semantic information. Second, SLAM-based methods are inherently less adaptive to posterior information, which causes them not generalizing well in complicated environments, especially in indoor scenes. Last but not least, SLAM-based methods are not specially designed for searching objects in unknown environments, which requires keeping balance between \textit{exploitation} (depth-first searching) and \textit{exploration} (regretting).

Recently, learning-based methods emerge and show powerful capabilities of solving complicated tasks. However, these methods generally have problems of interpretability and generalization, especially in the TDN task which require robots to operate in unseen domain. We argue that it is more natural and empirical to introduce \textit{a priori}\cite{chatzilygeroudis2019survey}\cite{liu2020efficient} to the learning model instead of training from scratch, considering how human teach ignorant babies. Introducing \textit{a priori} enables algorithms to achieve higher data efficiency, better model interpretability, and generalization. In indoor TDN tasks, one of the most useful prior information is the relationship among objects and rooms of different categories. Some recent works reason about the target direction using object relationships as \textit{a priori} in single-room environments\cite{ScenePriors, ORG, MJOLNIR}. However, common domestic scenes are composed of multiple rooms, thus more prior information such as room connection, object-in-room membership, and other implicitly structured relationships could be exploited, which are typically ignored in these works.

\begin{figure*}[ht]
	\centerline{\includegraphics[width=0.95\linewidth]{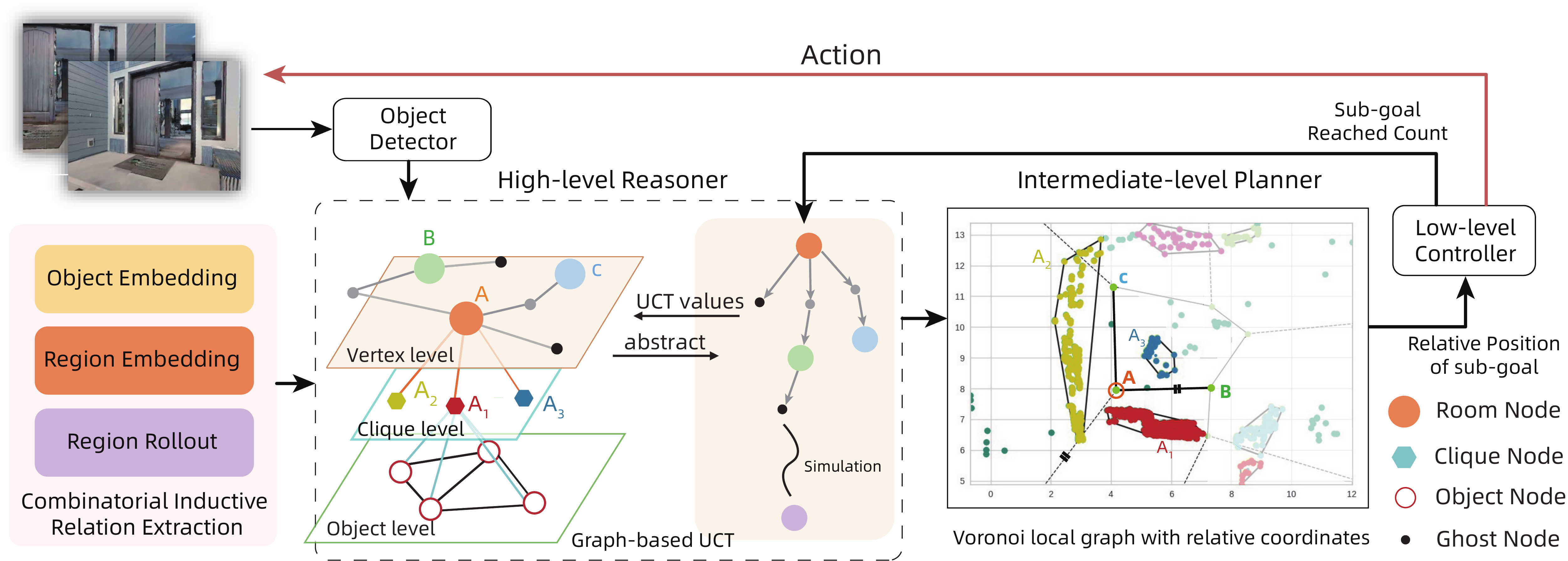}}
	\caption{The main hierarchical framework of ReVoLT method, which contains a high-level reasoner (infers semantic sub-goals), an intermediate-level planner (computes spatial location sub-goal), and a low-level controller (computes actions). The combinatorial relation extraction module provides a priori of the exploration value about the observed objects and regions through embedding similarity. Especially, Region Rollout model provides Monte Carlo simulation for UCT in a conditional GraphRNN (c-GraphRNN) way.}
\label{Hierarchical Framework}
\end{figure*}

In this paper, we propose a hierarchical navigation framework, \textit{Relational Reasoning and Voronoi Local graph planning} \textit{(ReVoLT)}, which comprises a combinatorial graph neural network for multiform domestic relations extraction, an UCT-based reasoning exploration, and an online Voronoi local graph for the semantic-spatial transition. The TDN task is concisely decomposed, allowing for separate and special designs for different modules, instead of operating in a mixed-up end-to-end manner. The detailed contributions are as follows:
	\begin{itemize}
	\item To extract multiform structural relations for reasoning, we propose combining unsupervised GraphSAGE \cite{GraphSAGE}, self-supervised GCN, and c-GraphRNN methods for learning object embedding, region embedding, and region rollout, respectively. 
	\item Based on the relation priors, the high-level reasoner (semantic reasoning) is abstracted as a bandit problem and adopts UCT to balance \textit{exploitation} (depth-first searching) and \textit{exploration} (regretting).
	\item Voronoi local graphs are constructed online for converting semantic sub-goals to spatial locations.
	\item It is found in the test results that the proposed framework is superior to state-of-the-art methods and achieves a higher success rate and success weighted by path length (SPL) with good generalization. 
	\end{itemize}

\section{Related Works}
Graph is useful for constructing relationships between non-Euclidean data, which has recently become popular for representing the physical structure of redundant robots to learn coordination between their multiple degrees of freedom \cite{liu2022robot}, learning traffic signal co-regulation between multi-junction networks \cite{liu2021learning}, and representing the state change of deformable and soft objects. It is also reasonable and natural to use graphs for relational reasoning to solve TDN problems. They have the advantage of replacing an explicit metric map like SLAM-based methods, inferring the approximate position of the target object based on observed objects. Most of these methods use GNNs to learn object-object proximity relationships but ignore the relationship between regions/rooms, thus it limits their task scenarios to a single room (using AI2Thor data set \cite{AI2Thor} in simulation for training). 
For example, Yang et al. \cite{ScenePriors} propose to use Graph Convolutional Network (GCN) to incorporate the prior knowledge about object relationship into a Deep Reinforcement Learning (DRL) framework as part of joint embedding. Their priors are obtained from large-scale scene understanding datasets and updated according to the current observation.

For navigation tasks in houses with multiple rooms, it is necessary to first reach the room that may contain the target object (e.g. refrigerator-kitchen), then search the target in object cliques. Therefore, the learning of prior knowledge should consider more relationships, including room-to-room connection and object-in-room membership. Wu et al. \cite{BRM} propose a memory structure based on the Bayesian graph model. It uses the probability relationship graph to get the prior house layout from the training set and estimates its posterior in the test set. However, this work does not combine object-level reasoning to achieve a complete TDN task. Chaplot et al. \cite{NTS} build a topological representation with associated semantic features and learn a prior semantic score function to evaluate the probability of potential nodes in a graph with various directions. However, they provide target images,
 which is impractical in domestic scenarios, while our method only uses target labels. They subsequently extend the Active Neural SLAM system \cite{activeneuralslam}, to learn semantic priors using a semantically aware long-term policy for label target navigation task \cite{SemExp} and won CVPR 2020 Habitat ObjectNav Challenge\footnote{https://aihabitat.org/challenge/2020/} \cite{objectnav}. It is worth mentioning that they also point out the end-to-end learning-based methods suffer from large sample complexity and poor generalization as they memorize object locations and appearance in training environments\cite{SemExp}, which prompt us to consider the hierarchical framework with a topological graph. Table \ref{related_work} only lists TDN methods with label target and relational reasoning.

\begin{table} 
	\centering
	\caption{Performance of existing TDN methods with\\ various experiment setting} 
	\label{related_work}
	\begin{threeparttable}
	\setlength{\tabcolsep}{2.5mm}{
	\begin{tabular}{ccccc}
		\toprule
		Method& Room Scale &  Dataset & SR(\%) & SPL(\%)\\
		\midrule
		Scene-prior \cite{ScenePriors} & Single & AI2-THOR & 35.4 &  10.9 \\
		SAVN \cite{SAVN} & Single & AI2-THOR & 35.7 &  9.3\\
		MJOLNIR \cite{MJOLNIR} & Single & AI2-THOR & 65.3 & 21.1\\
		BRM \cite{BRM} & Multiple & House3D & - & - \\
		SemExp$^\dagger$ \cite{SemExp} & Multiple & Matterport3D & 36.0 & 14.4 \\
		\bottomrule[1pt]
	\end{tabular}
	}	
	 	\begin{tablenotes}
			\footnotesize
			\item[$ \dagger $] SemExp won the first place in CVPR Habitat 2020 competition.
		\end{tablenotes}
	\end{threeparttable}
\end{table}

\begin{figure*}[ht]
	\centerline{\includegraphics[width=1\linewidth]{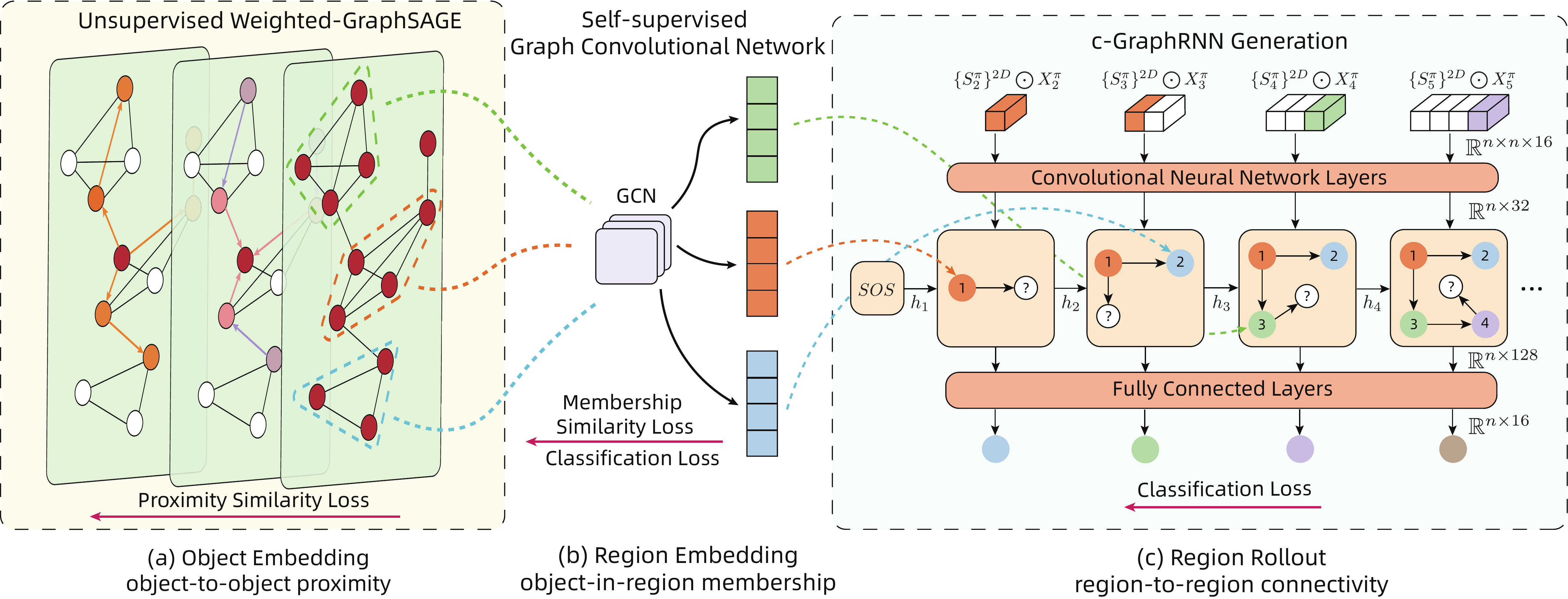}}
	\caption{Combinatorial relation extraction module. (a) Obtain object embedding via unsupervised weighted-GraphSAGE; (b) Region embedding is received by passing a sub-graph with object embedding to GCN layers; (c) According to the house structure of region connectivity, a GraphRNN-based model is used to learn the structure distribution and generate possible feature of future regions node by node.}
	\label{Inductive}
\end{figure*}

\section{Revolt Reasoning \& Planning with Hierarchical Framework}
This task needs to be re-examined from the perspective of bionics. Imagine a human facing such a task when he enters an unknown house. He will not feel confused due to the prior knowledge about domestic scenes he has. It is natural for us to first roughly determine the type of room based on categories of multiple observed objects in the current room (e.g. a bedroom). According to the object-in-room membership, the exploration value $\mathbf{V}(t|cur\_room)$ of the target object $t$ in the current room can be obtained. At the same time, some potential but unexplored passages (e.g. a door or hallway) can be determined as \textit{ghost} nodes like \cite{NTS}. The structural relationship of the house layout and room connection can help us predict categories and value $\mathbf{V}(t|next\_room)$ of next rooms connected by ghost nodes.

Except for these priors, dynamic decisions also should be made in a specific task, rather than just applying experience mechanically. Reasoning procedure which contains intelligent exploration and exploitation is one of the winning strategies. Thus, our approach focuses on solving the following two problems:
	\begin{itemize}
	\item How to obtain a more effective prior conditional exploration value in a structured form?
	\item How to make efficient decisions between multiple feasible paths based on exploration values? 
	\end{itemize}

The remainder of this section is organized as follows. In
subsection \ref{3A}, \ref{3B}, \ref{3C}, we present a combinatorial relation extraction module (Fig. \ref{Inductive}) using GNNs, which learns three different relationships in a unified paradigm. A UCT-based online reasoner is described in subsection \ref{3D}. In \ref{3E}, we consider the coarse spatial information and build an intermediate-level planner through online Voronoi construction. Finally, the whole \textit{ReVoLT} hierarchical framework is summarized in subsection \ref{3F} (Fig. \ref{Hierarchical Framework}).

\subsection{Object Embedding learning}\label{3A}
As illustrated in Fig. \ref{Inductive} (a), the object-to-object relationship consists of not only pair-wise semantic similarity, but also distances and the number of hops between object pairs. We first extract an object-level graph $\mathcal{G}_o(\mathcal{V}_o,\mathcal{E}_o)$ through object positions $pos$ and category $\mathcal{C}_o$ from Matterport3D dataset. Objects in the same room are fully connected. As for object pairs in different rooms, only those closest to a common door have an connecting edge. This is useful for the robot to infer objects that are strongly related to the target just using object-level embedding.

GraphSAGE \cite{GraphSAGE} is a popular model in the node embedding field. We adopt it to obtain the embedding of each object category to fuse semantics and proximity relationships with other categories. Our node embedding procedure uses GloVe \cite{Glove} as the initial node semantic feature $\left\{\mathbf{x}_{v}, \forall v \in \mathcal{V}_o\right\}$, and employ an unsupervised form of GraphSAGE with a loss that penalizes the embedding similarity between two objects far apart and reward the adjacent two. Different from the original GraphSAGE, edge features $\left\{\omega_{e: u\rightarrow v}, \forall e \in \mathcal{E}_o\right\}$ are also taken into account in aggregation and loss calculations. For each search depth $k$, weight matrices $\mathbf{W}^k,  \forall k\in \{1,\dots,K\}$, we employ an edge-weighted mean aggregator which simply takes the element-wise mean of the vectors in $\{h^{k-1}_u, \forall u\in \mathcal{N}(v)\}$ to aggregate information from node neighbors:
\begin{equation}
	\begin{aligned}
		&\mathbf{h}_{v}^{0}  \leftarrow x_v, \forall v\in \mathcal{V}\\
		\mathbf{h}_{v}^{k} \leftarrow &\sigma\left(\mathbf{W}^k \cdot \text{mean}(\{\mathbf{h}_{v}^{k-1}\} \cup\{\omega_{u\to v}\cdot\mathbf{h}_{u}^{k-1}\})\right)
	\end{aligned}
\end{equation}

Then an edge-weighted loss function is applied to the output $\{\mathbf{z}_v, \forall v\in \mathcal{V}_o\}$, and tune the weight matrices $\mathbf{W}^k$:
\begin{equation}
	\begin{aligned}
		\mathcal{L}_{\mathcal{G}_o}\left(\mathbf{z}_{v}\right)=&-\log \left(\sigma\left(\omega_{u\to v}\mathbf{z}_{v}^{\top} \mathbf{z}_{u}\right)\right)\\&-Q \cdot \mathbb{E}_{u_{n} \sim P_{n}(v)} \log \left(\sigma\left(-\omega_{u\to v}\mathbf{z}_{v}^{\top} \mathbf{z}_{u_{n}}\right)\right)
	\end{aligned}
\end{equation}
where $P_n$ is a negative sampling distribution, $Q$ defines the number of negative samples, $\sigma$ is the sigmoid function.

Since object embeddings with the same category $\{\mathbf{z}_c, \forall c \in \mathcal{C}_o\}$ should have consistent representation, another mean aggregation is performed on the embeddings of same category between the final GraphSAGE aggregation and loss function. This overwrites the original value with the final embedding  for each category $\{\mathbf{z}_{c} \leftarrow \text{mean}(\mathbf{h}_v^K), \textrm{if } \mathcal{C}_o(v) =c\}$.

\subsection{Region Embedding learning}\label{3B}

Apart from the pairwise relationship between objects, the many-to-one relationship between an object and a room or region is also indispensable for inferring the existence possibility of the target object in a certain room or among multiple observed objects. Besides, to evaluate the similarity, relationships of different levels should have a unified paradigm to obtain representation of consistent metrics. Therefore, for region-level sub-graphs, we still opt for the same embedding representation procedure. This part is shown in Fig. \ref{Inductive} (b).

Region embedding is carried out in a self-supervised form. We take the sub-graph $\mathcal{G}_r(\mathcal{V}_r, \mathcal{E}_r)$ as input, with embedding of objects in the same region $\{\mathbf{z}_c, \forall c \in \mathcal{C}_o\}$ as nodes and weighted spatial distances as edges. The batch composed of these sub-graphs is passed into the GCN\cite{GCN}, and the corresponding region embedding $\{\mathbf{r}_{v}, \forall v \in \mathcal{V}_r\}$ is obtained. Similarly from the previous procedure, for region embedding with the same label, a mean aggregation is performed to obtain a uniform vector representation $\{\mathbf{r}_{l}, \forall l \in \mathcal{L}_r\}$. Since there is no need to do multi-hop aggregations at region-level, a simple GCN layer is applied rather than GraphSAGE. 

To enable membership calculation between region embedding $\mathbf{r}_{l}$ and object embedding $\mathbf{z}_c$ and distinguish regions with different labels, we use a combined loss which comprises two parts: the classification loss of embedding label and the membership loss of object-in-region:
\begin{equation}
	\begin{aligned}
		\mathcal{L}_{\mathcal{G}_r}\left(\mathbf{r}_{v}\right)=&-\log \left(\sigma\left(\mathbf{r}_{v}^{\top} \mathbf{z}_{u}\right)\right)
		\\&-Q \cdot \mathbb{E}_{u_{n} \sim P_{n}(v)} \log \left(\sigma\left(-\mathbf{r}_{v}^{\top} \mathbf{z}_{u_{n}}\right)\right)
		\\&- \dfrac{1}{n}\sum_{i=1}^n l_v\log(\hat{l}(\mathbf{r}_v))
	\end{aligned}
\end{equation}
where $P_n(v)$ represents objects not in region $v$, and $\hat{l}(\cdot)$ is a multi-layer perceptron (MLP) network.

\subsection{Region Rollout learning}\label{3C}

As the third and most important part of relation extraction, the structural relationship reasoning ability plays a crucial role in understanding the correct direction of navigation and shortening the exploration period. To achieve this, the joint probability $p(\mathcal{G}_h)$ of houses need to be learned to conceive a probable house layout memory $\mathcal{G}_h\sim p(\mathcal{G}_h|\mathcal{G}_{sub})$ conditioned on observed regions $\mathcal{G}_{sub}$. However, its sample space might not be easily characterized. Thus, the house graphs are modeled as sequences by following the idea of GraphRNN\cite{GraphRNN}, and redefine some concepts to make it more suitable for conditional reasoning with embedding. This part is shown in Fig. \ref{Inductive} (c).
\begin{equation}
	\begin{aligned}
	S^{\pi}=f_{s}(\mathcal{G}_h, \pi)=\left(A_{1}^{\pi}, \ldots, A_{n}^{\pi}\right)
	\end{aligned}
\end{equation}
where $\pi$ represents the node order, and each element $A_{i}^{\pi} \in\{0,1\}_{(i-1)\times(i-1)}, i \in\{1, \ldots, n\}$ is an adjacent matrix referring the edges between node $\pi(v_i)$ and its previous nodes $\pi(v_j), j\in\{1, \dots, i-1\}$ already in the graph.

Since each $A_{i}^{\pi}$ has variable dimensions, we first fill them up to the maximum dimension $n$ and then repeat the 2D matrix 16 times to form a 3D matrix with $n\times n\times 16$ dimensions as an edge mask where 16 is the embedding length. Therefore, a featured graph can be expressed as the element-wise product of the region embedding matrix $X^\pi$ under corresponding order and sequence matrix $\{S^\pi\}^{3D}$: 
\begin{equation}
	\begin{aligned}
	p(\mathcal{G})=\prod_{i=1}^{n+1} p\left(\mathbf{x}_{i}^{\pi} \mid (\{S_{1}^{\pi}\}^{3D}, \ldots, \{S_{i-1}^{\pi}\}^{3D})\odot X_{i-1}^{\pi}\right)
	\end{aligned}
\end{equation}
where $X_{i-1}^{\pi}$ is the embedding matrix with $(i-1)\times(i-1)\times16$ dimensions referring to region embeddings before region $\pi(v_i)$, and $\mathbf{x}_{i}^{\pi}$ refers to the embedding of $\pi(v_i)$.

Passing $\{S^{\pi}\}^{3D} \odot X^\pi$ as a sequence into GRU or LSTM, we can get the structure distribution of houses. This allows us to predict the next region embedding and label under the condition of the observed subgraph. The loss function of the Region Rollout network is a CrossEntropy between generated embedding label and the real label:
\begin{equation}
	\begin{aligned}
	\mathcal{L}_{\mathcal{G}_h}(\mathbf{x}^\pi_i) =  -\dfrac{1}{n}\sum_{i=1}^n l_i \text{ log-softmax}[(\mathbf{x}^\pi_i)^T\mathbf{r}_j], \forall j \in \mathcal{L}_r
	\end{aligned}
\end{equation}

In conclusion, with the combination of \ref{3A} unsupervised edge-weighted GraphSAGE object embedding learning, \ref{3B} self-supervised GCN region embedding learning, and \ref{3C} c-GraphRNN conditional region rollout, we can now extract multiform structural relationships. Meanwhile, embedding is used as a unified paradigm for representation, and the similarity between objects or regions (either observed or predicted) embeddings and the target object embedding is used as a prior to guide the exploration in an unknown domain. 

\subsection{Reasoning and Exploring as a Bandit Problem}\label{3D}
\begin{figure}[ht]
	\centerline{\includegraphics[width=1\linewidth]{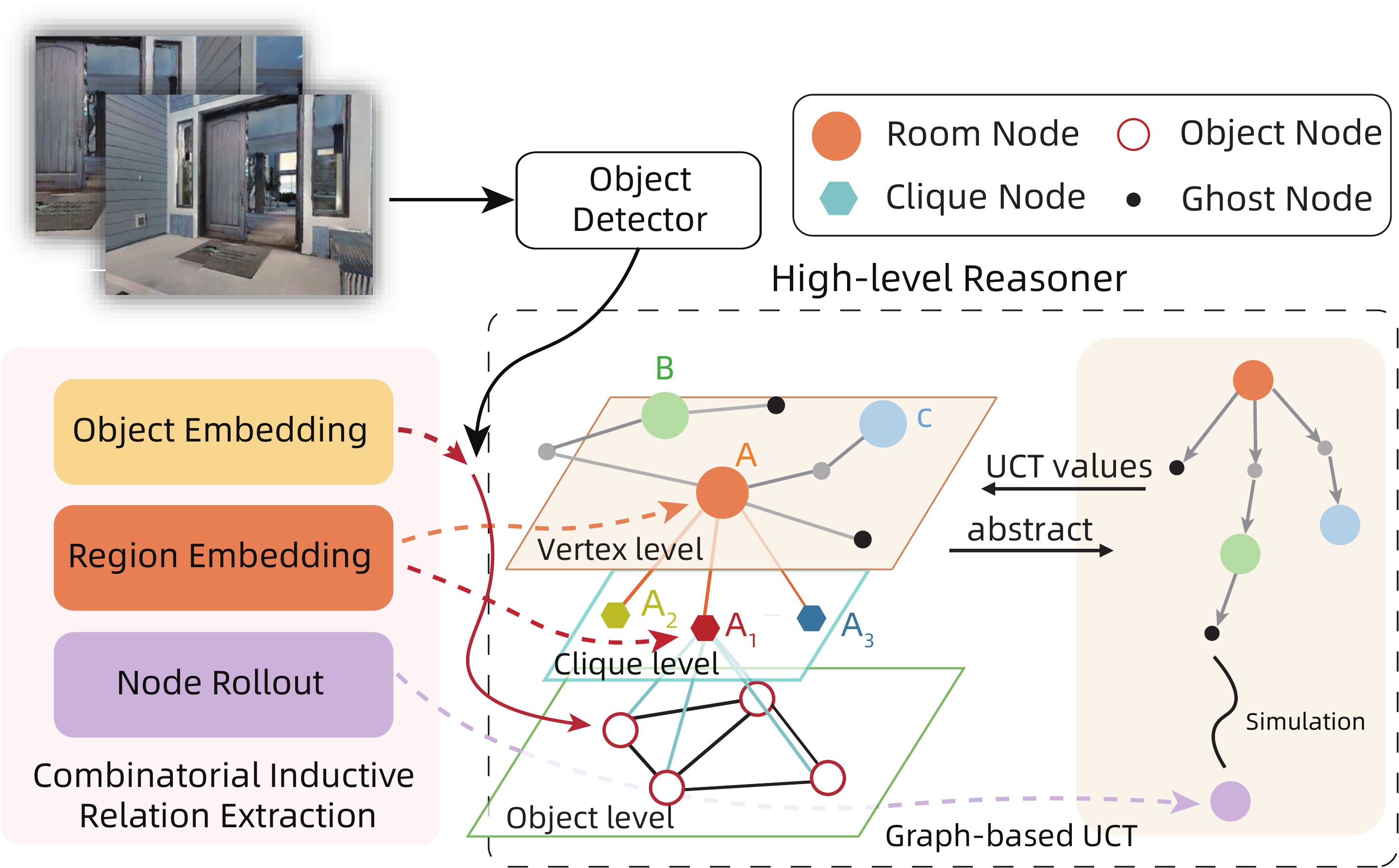}}
	\caption{In a specific task, a multi-layer topological graph is constructed based on visual front-end, and a tree with the birthplace as the root node is abstracted from the graph. The \textit{clique} refers to a collection of adjacent objects or a bunch of non-semantic obstacles, and the \textit{vertex} refers to an observed navigable location. Each gray \textit{ghost} node has connected two vertices, and only stores the relative position of the connected vertices to assist localization, without being used as a navigation sub-goal. The black \textit{ghost} nodes refer to unknown areas and promote exploration.}
	\label{graph&UCT}
\end{figure}

A prior alone cannot lead to success. Inspired by \cite{NTS}, a posterior topological representation is also constructed in each specific task to combine experience with practice.  Specifically, we build a multi-layer posterior topological graph covering all object-level, clique-level and vertex-level. 
\textit{clique} divides rooms into small clustered regions and reduces the burden of the visual front-end. Each \textit{vertex} governs the three nearest cliques. Object Embedding network provides the object node features, and Region Embedding network generates the features of both clique and vertex from their attached objects. Region Rollout network gives an evaluation about \textit{ghost} nodes. However, there are always situations contrary to experience in reality. In other words, robots must have the ability to balance exploration and exploitation online.
We adopt Upper Confidence Bound for Tree (UCT) method\cite{bandittree} to set an online bonus. The simulation procedure of UCT is supported by the Region Rollout network, thus the robot is not only able to obtain the bonus from reached count, but also estimate the future exploration value inductive bias $\omega_i$ of selected path. It can effectively prevent the robot from being trapped in a useless area. The combined effect of inductive bias $\omega$ and bonus will discourage the repetitive search near negative (non-success) sub-goals and drive the robot to return to parent nodes for back-tracking, which we term \textit{Revolt Reasoning}. The word \textit{Revolt} summarizes the characteristics of our method vividly, which allows robots to regret at nodes with low exploration value, discarding them and returning to previous paths. To avoid robots wandering between two goals, it is necessary to introduce a navigation loss term $\mathcal{L}_{dis}$ to penalize node distances. Hence, we can finally obtain the exploration value $\mathbf{V}$ of the node $i$ as:
\begin{equation}
	\begin{aligned}
		\mathbf{V}(t\vert i) = \dfrac{\Sigma^m_{i\rightarrow j} \omega_j}{m}  + c_1\sqrt{\dfrac{\ln N_i}{n_i}} -c_2\mathcal{L}_{dis}
	\end{aligned}
\end{equation}
where factors $c_1$ and $c_2$ are set as 1 and 0.5. $j$ refers to one of node $i$'s descendants in the tree, and $m$ is its total number. $N_i$ is the total arrivals of node $i$ and its descendants, while $n_i$ just represents arrivals of node $i$.

\subsection{Online constructed Voronoi local graph}\label{3E}
\begin{figure}[ht]
	\centerline{\includegraphics[width=1\linewidth]{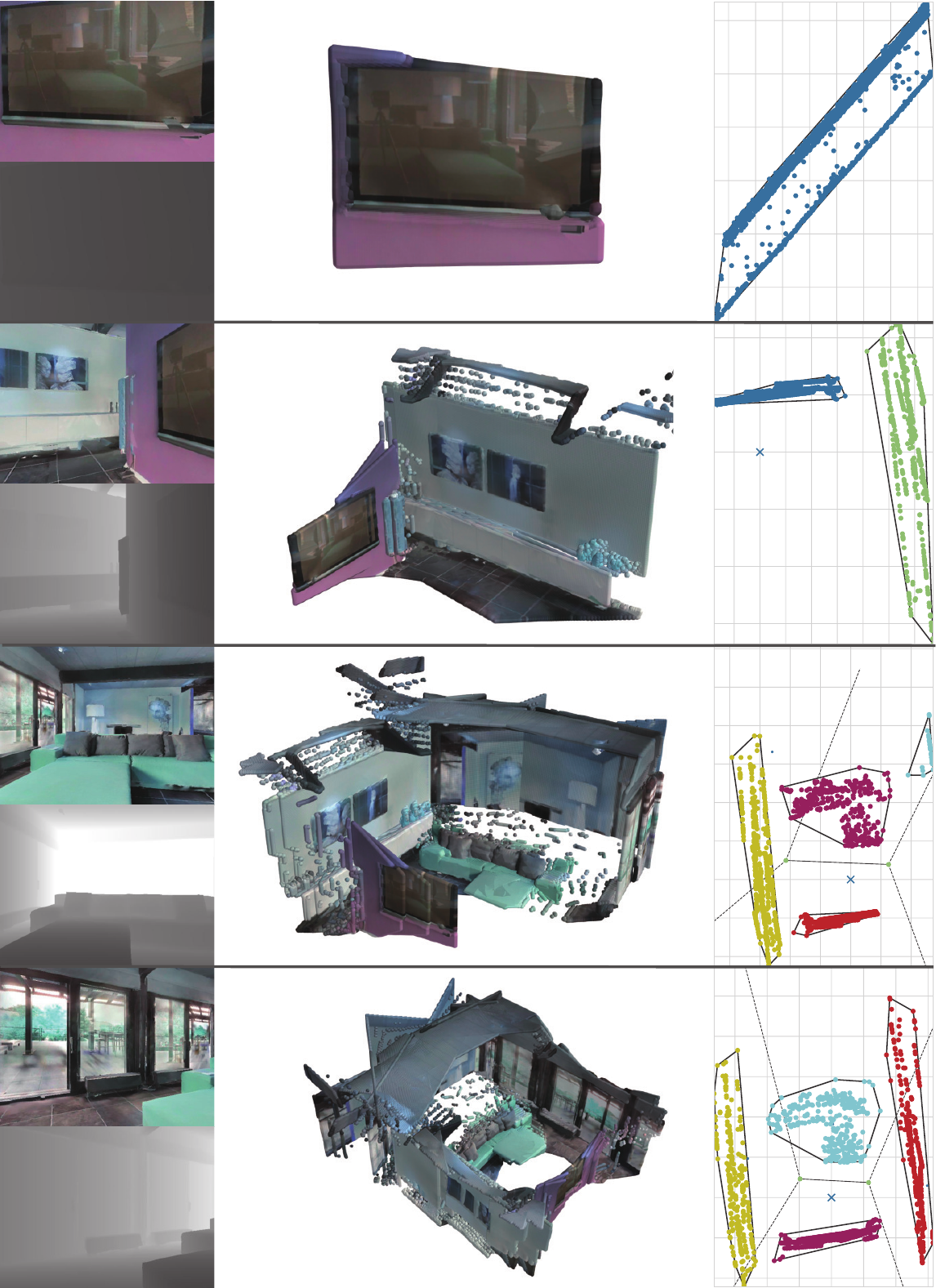}} 
	\caption{Combining the depth information with robot's pose in a short period, then we can get a simple 3D reconstruction result. A Voronoi local graph can be constructed through DBSCAN clustering after projecting the 3D map as a 2D obstacle scatter plot.}
	\label{reconstruction}
\end{figure}
The reasoner only gives a semantic node id in a graph as a sub-goal. If the low-level controller directly uses it as a navigation goal, it will inevitably lead to over-coupling and increase the difficulty of navigation success. We can refer to the hierarchical human central nervous system composed of the brain, \textit{cerebellum}, \textit{brain-stem} and \textit{spinal cord}\cite{purves2008cognitive}, if the high-level reasoner is compared to the \textit{brain}, then the \textit{skeletal muscle} is the low-level motor controller. The brain does not directly transmit motion instructions to the \textit{skeletal muscles}, but passes it through the \textit{brain-stem}, \textit{spinal cord} and other lower-level central nervous system for information conversion\cite{spinal_cord}. Besides, the \textit{brain} does not actually support high-speed, low-latency information interaction while controlling a motion\cite{brainslow}. Therefore, it is necessary to use a RGB-D camera and an odometer to construct a local Voronoi graph, offering approximate relative coordinates of the sub-goal within a reachable range as an input to the low-level controller. The Voronoi graph can record the relationship between the robot and obstacles, and provide an available path. Since the TDN task is map-less, we construct a local Voronoi graph within a fixed step online.

Conditioning on the depth information, the parameters (internal and external) of the camera, and the odometer information, obstacles in depth images can be easily converted into coordinates in a world coordinate system. This system is derived from the birth pose of the robot. Projecting this partially reconstructed 3D map onto a 2D plane along the vertical axis forms a scatter diagram depicting obstacles. We can construct a Voronoi diagram online by segmenting navigable paths and explorable \textit{cliques} with multiple related objects. Different from traditional methods \cite{traditionalVoronoi}, we use DBSCAN \cite{DBSCAN1, DBSCAN2} (a density-based clustering algorithm) to cluster the scattered points of adjacent obstacles into convex hulls first, and then filter out noise points. Followed by constructing Delaunay triangle with the center of scattered points in the convex hull, thereby generating a Voronoi diagram.

\subsection{Hierarchical reasoning and planning for navigation}\label{3F}
\begin{figure}[ht]
	\centerline{\includegraphics[width=1\linewidth]{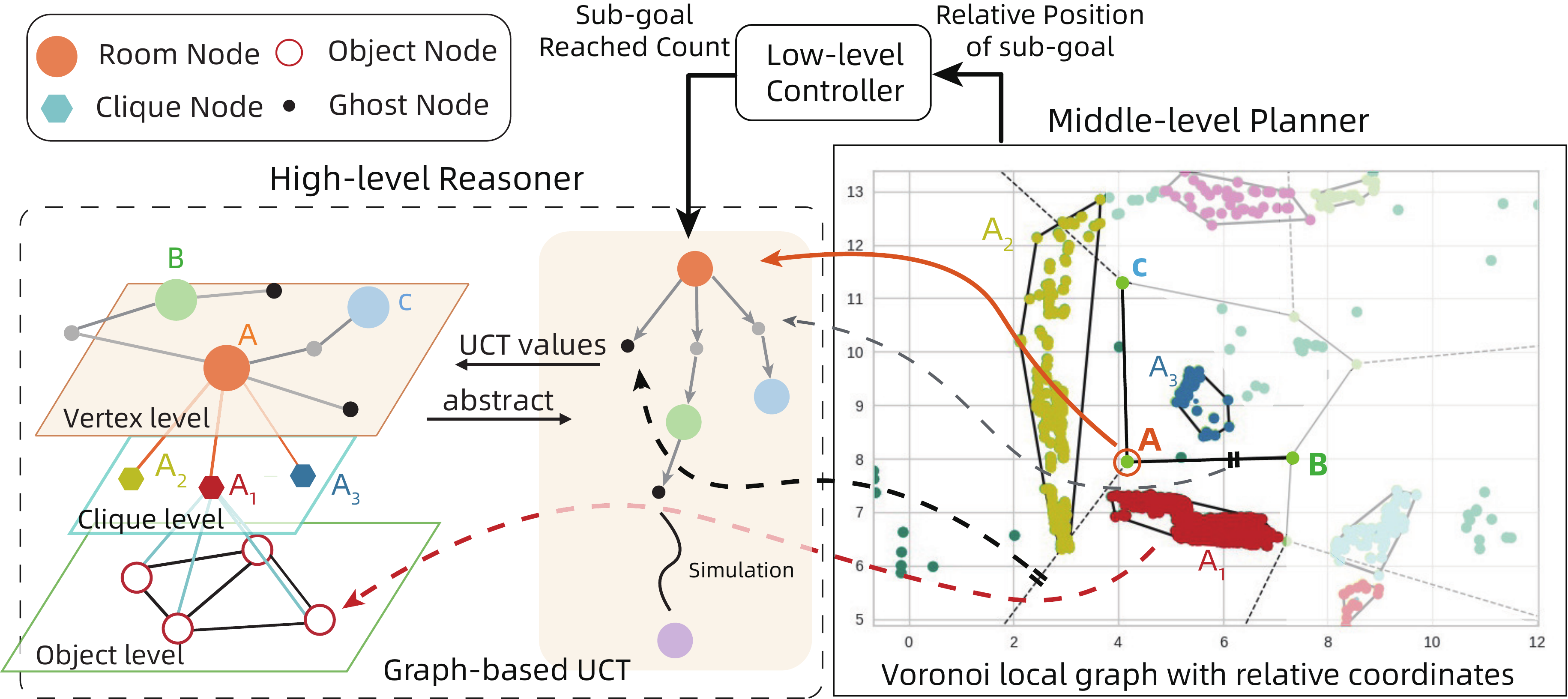}}
	\caption{The semantic sub-goal is converted into relative coordinates by the Voronoi-based intermediate-level planner.}
	\label{Hierarchical Voronoi}
\end{figure}

In this section, we will summarize how the proposed reasoner and planner cooperate to complete navigation tasks. The curves in Fig. \ref{Hierarchical Voronoi} show the correspondence of concepts between the topological graph in \textit{reasoner} and the Voronoi diagram in \textit{planner}. The aggregation of obstacles is regarded as a \textit{clique}, each of which attaches and records all \textit{objects} in its convex hull, and evaluates its inductive bias value according to the object-in-region membership via the Region Embedding network. The position of a \textit{vertex} is generated by Voronoi. Multiple \textit{cliques} and their subordinate \textit{objects} surrounding the \textit{vertex} jointly determine the general room label of it, and use the label for the inductive bias evaluation. Relative directions and distances between two adjacent vertex nodes are stored in gray \textit{ghost} nodes. Since the robot exploits relative coordinates and directions, it effectively avoids the influence of odometer and depth camera error, thus insensitive to cumulative error. Besides, thanks to the Voronoi local diagram, only short-period scatter data need to be saved, and there is no need to consider the closed-loop matching problem like SLAM.


With the construction of Voronoi diagram and its transformation to a hierarchical topology, we can conduct reasoning in \textit{vertex/clique}-level and \textit{object}-level, searching for the best \textit{vertex} position and the most likely \textit{clique} based on the exploration value. After selecting a \textit{clique}, the robot will navigate towards it, and explore it more explicitly as \textit{object}-level reasoning. Besides, the Voronoi diagram provides the evidence for choosing the next best view of one \textit{clique}. By changing multiple perspectives, the robot can find the target object in a clique more efficiently.

\section{Experiments}
\subsection{Experiment Setup}
We use the Habitat simulator \cite{habitat19iccv} with Matterport3D \cite{Matterport3D} environment as our experiment platform. Habitat simulator is a 3D simulator with configurable agents, multiple sensors, and generic 3D dataset handling. Matterport3D dataset contains 90 houses with 40 categories of objects and 31 labels of regions. It also provides detailed object and region segmentation information. Here we just focus on 21 categories of target object required by the task: 
\textit{chair, table, picture, cabinet, cushion, sofa, bed, chest of drawers, plant, sink, toilet, stool, towel, tv monitor, shower, bathtub, counter, fireplace, gym equipment, seating, clothes} and also ignore some meaningless room labels, like \textit{outdoor}, \textit{no label}, \textit{other room} and \textit{empty room}. We use YOLOv4 \cite{yolov4} as our object detection module, which is fine-tuned using objects in Matterport3D dataset. Because the aiming of low-level controller is the same as PointNav task's \cite{PointNav}, we adapt a pre-trained state-of-the-art PointNav method \textit{occupancy anticipation} \cite{occupancy} as our controller. 

During a specific TDN task, the robot is spawned at a random location in a certain house and is demanded to find a object of a given category as quickly as possible. The task is evaluated with three commonly used indicators: \textbf{Success Rate (SR)}, the \textbf{Success weighted by Path Length (SPL)} and \textbf{Distance to Success (DTS)}. SR represents the number of times the target was found in multiple episodes and is defined as $\frac{1}{N}\sum^N_{i=1}Su_i$, where $N$ is the number of total episodes and $Su_i$ is a binary value representing the success or failure of the $i$-th episode. SPL depicts both success and the optimal path length, it is defined as $\frac{1}{N} \sum_{i=1}^{N} S_{i} \frac{L_{i}}{\max \left(P_{i}, L_{i}\right)}$. Here we use the shortest length provided by the simulator as $L_{i}$ and the path length of the robot as $P_i$ in episode $i$. DTS is the distance of the agent from the success threshold boundary when the episode ends. The boundary is set to $1m$ and the maximum episode length is $500$ steps, which are the same as \cite{SemExp}.


Furthermore, our navigation task has two modes: \textit{independent (ReVoLT-i)} and \textit{continuous (ReVoLT-c)}. Independent mode is the traditional one, the environment is reset after each episode and the robot clears its memory. While the continuous mode allows the robot to keep the topological graph if it resets in the same house. It is used for evaluating the robot's capability of keeping and updating the environment memory.

\subsection{Baselines}
\textbf{Random}: At each step, the agent randomly samples an action from the action space with a uniform distribution.

\textbf{RGBD + DD-PPO}: This baseline is provided by ObjectNav Challenge 2020 \cite{habitat19iccv}. Directly pass RGB-D information to an end-to-end DD-PPO and output an action from the policy.

\textbf{Active Neural SLAM}: This baseline uses an exploration policy trained to maximize coverage from \cite{activeneuralslam}, followed by the heuristic-based local policy as described above.

\textbf{SemExp}: Since \cite{SemExp} has not open-sourced their code, we directly use results in their paper as a state-of-the-art method.

\subsection{Results}
\subsubsection{results of combinatorial relation embeddings}
The Object Embedding network obtains classification accuracy of 91\%. The Region Embedding network obtains  membership accuracy of 78\% and classification accuracy of 75\%. The Region Rollout network reaches prediction accuracy of 45\% in the test set, which is acceptable since room relationships are not significant inherently. 
\subsubsection{results of the whole TDN task}
\begin{figure}[h]
	\centerline{\includegraphics[width=1\linewidth]{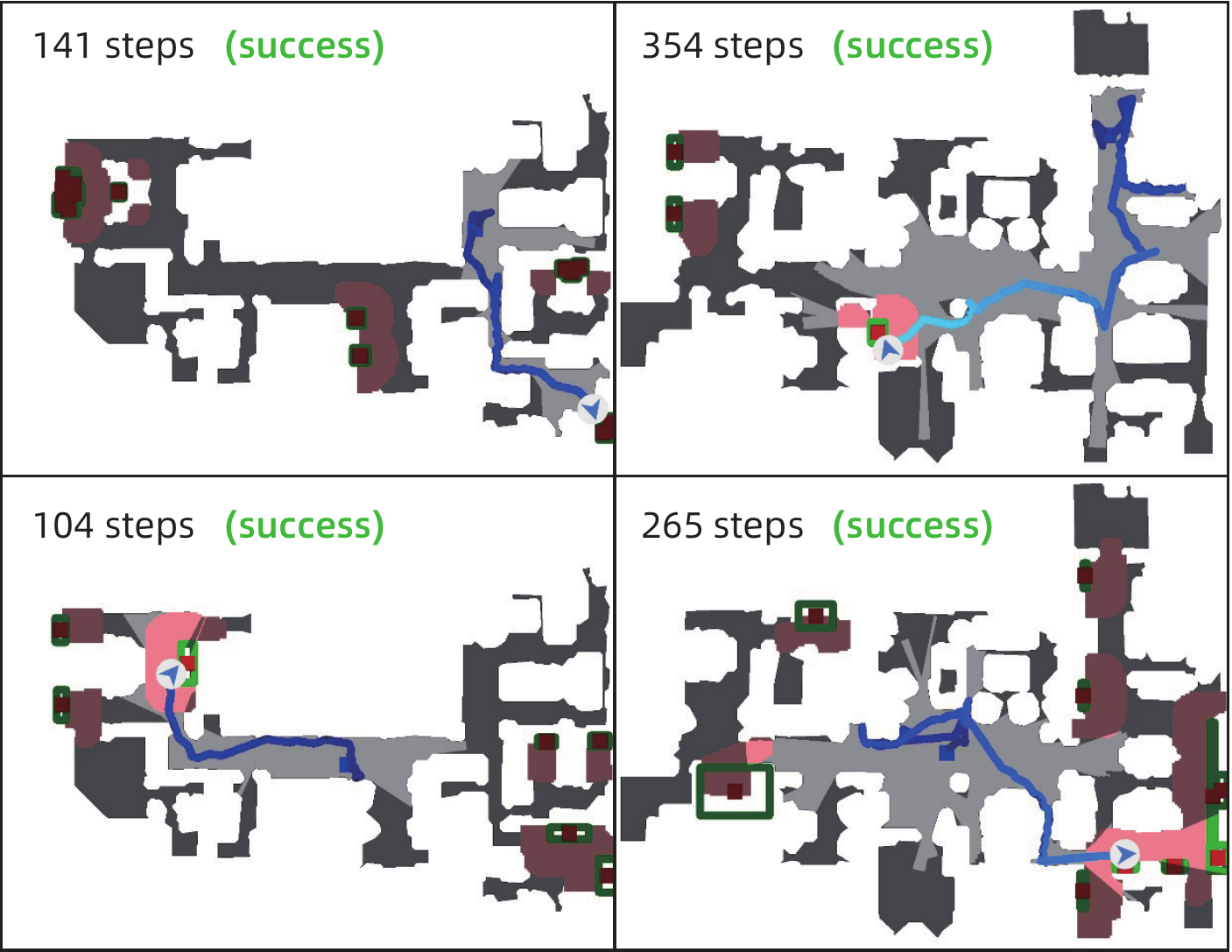}}
	\caption{Top-down maps of four successful tasks while using \textit{ReVoLT-i}. The blue squares are the beginning positions, the blue curves are the robot trajectories, and arrows represent the robot's current positions. Targets are highlighted with green boxes, and pink areas refer to the success threshold boundary. The color of the trajectory is a gradient from dark to light, and the brighter the end indicates the longer the path.}
	\label{Results}
\end{figure}
\begin{table}[ht]
	\centering
	\caption{Performance Comparison} 
	\label{Result_table}
	\begin{threeparttable}
		\setlength{\tabcolsep}{5mm}{
			\begin{tabular}{cccc}
				\toprule[1.5pt]
				Method& SR(\%) & SPL & DTS (m)\\
				\midrule
				Random & 0 &  0 & 10.3298\\
				RGBD + DD-PPO & 6.2 & 0.021 & 9.3162\\
				Active Neural SLAM & 32.1 & 0.119 & 7.056\\
				SemExp$^1$  & 36.0 & 0.144 & 6.733\\
				\midrule
				\textbf{\textit{ReVoLT-i small$^*$}} & \textbf{66.7}& \textbf{0.265} & \textbf{0.9762} \\
				\textbf{\textit{ReVoLT-i$^*$}} & \textbf{62.5}& \textbf{0.102} & \textbf{1.0511}\\
				\textbf{\textit{ReVoLT-c$^*$}} & \textbf{85.7}& \textbf{0.070} & \textbf{0.0253}\\
				\bottomrule[1.5pt]
			\end{tabular}
		}
		\begin{tablenotes}
			\footnotesize
			\item[1] The 1st prize of AI Habitat 2020
			\item[*] These three refer to \textbf{\textit{small}} mode with only $6$ categories target like SemExp, independence mode (\textbf{\textit{-i}}) and continuous mode (\textbf{\textit{-c}}) of \textit{ReVoLT}. 
		\end{tablenotes}
	\end{threeparttable}
\end{table}
The results of baseline methods and \textit{ReVoLT} is shown in Table \ref{Result_table}. It can be seen that both of our models significantly outperform the current state-of-the-art. \textit{ReVoLT-i small} has $\approx 80\%$  increase in SR and nearly twice than SemExp in SPL. This confirms our hypothesis that separating prior learning and control policy in a hierarchical framework is indeed a wise approach than directly learning a semantically-aware policy. Besides, the standard \textit{ReVoLT-i} with $19$ categories of targets still achieve a higher SR and SPL. By applying the continuous mode, the robot retains a memory belonging to the same house, which allows it to find observed targets with a higher SR. 

\section{Ablation Study}
\begin{figure}[ht]
	\centerline{\includegraphics[width=1\linewidth]{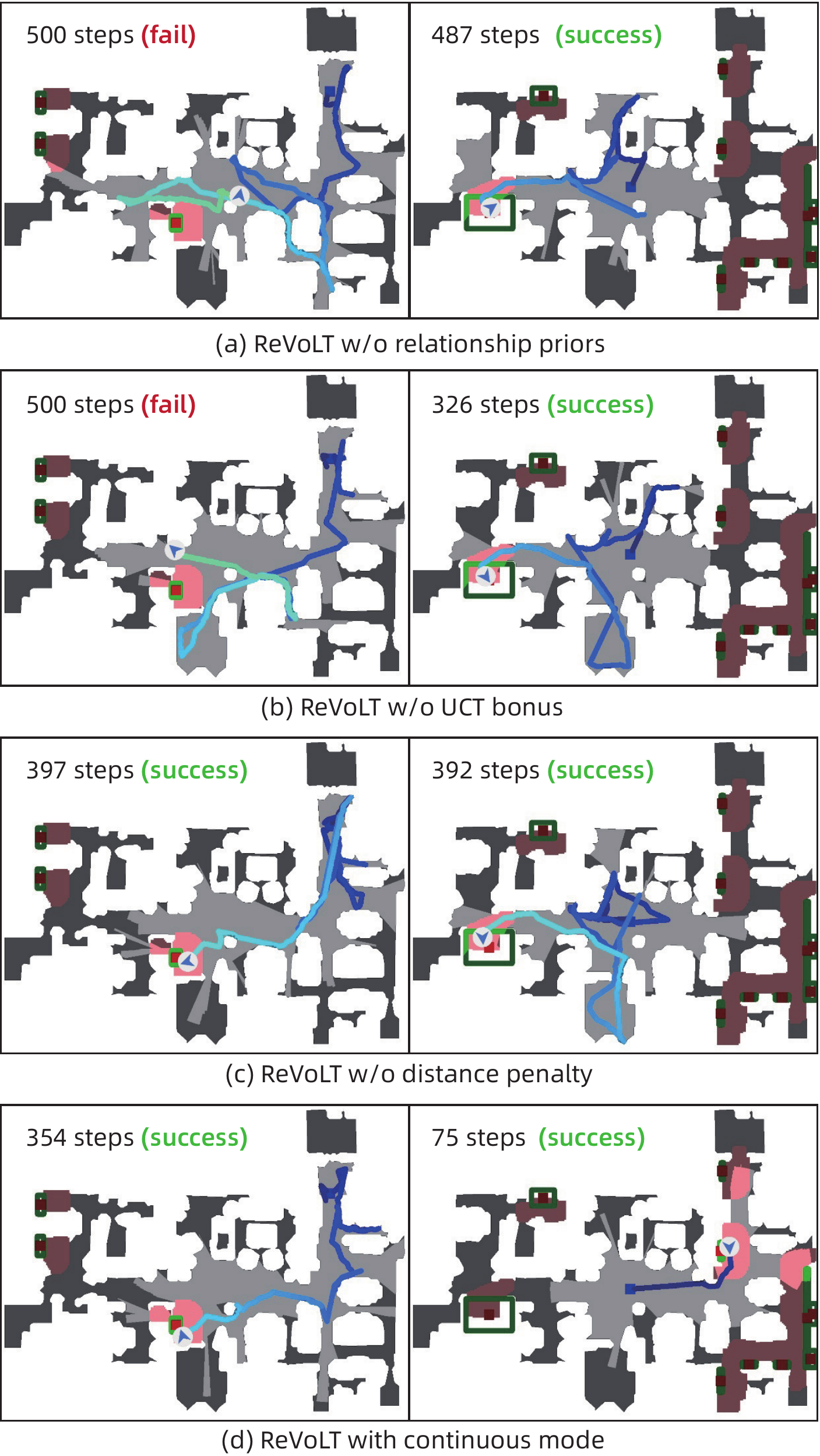}}
	\caption{In response to the three parts of exploration value function, we conduct ablation experiments respectively and illustrate them in top-down maps.}
	\label{Ablation}
\end{figure}

The success of \textit{ReVoLT} is attributed to the relationship priors provided by the combinatorial GNNs, the online bonus by UCT, and the distance penalty. Therefore, we set three extra experiments with the same Voronoi-based planner and low-level controller to reveal their impacts, respectively. Moreover, the results of the continuous mode are also presented below. The performance of all varieties is listed in Table \ref{Ablation_table}.\\
\begin{table}[ht]
	\centering
	\caption{Performance of Ablation Experiments}
	\label{Ablation_table}
	\begin{threeparttable}
		\setlength{\tabcolsep}{5mm}{
			\begin{tabular}{cccc}
				\toprule[1.5pt]
				Method& SR(\%) & SPL & DTS (m)\\
				\midrule
				\textit{ReVoLT-i} & 62.5 &  0.102 & 1.0511\\
				\textit{ReVoLT-c} & 85.7 &  0.070 &  0.0253 \\
				\textit{ReVoLT w/o priors} & 25.0 & 0.003 & 1.4129\\
				\textit{ReVoLT w/o bonus} & 60.0 & 0.034 & 0.8139\\
				\textit{ReVoLT w/o distance}  & 54.5 & 0.030 & 1.2689\\
				\bottomrule[1.5pt]
			\end{tabular}
		}
	\end{threeparttable}
\end{table} 
\textbf{ReVoLT w/o relationship priors.} Sub-goal in the navigation without priors can be generated according to the distance of the observed cliques. Compared to Fig. \ref{Ablation} (a) with Fig. \ref{Results}, we find that the lack of semantic relationship profoundly affects the robot's path decision, making it not interested in the region with a target even though it is just nearby. Besides, the lack of region classification and region rollout makes the robot unable to use the observed semantic information to reason about relationships, resulting in longer paths.\\
\textbf{ReVoLT w/o UCT bonus.} The bonus is replaced with a fixed threshold. If the robot reaches the same clique or vertex node more than twice, then this node will no longer be selected as a sub-goal. The corresponding top-down maps are illustrated in Fig. \ref{Ablation} (b). Without a UCT bonus, the robot falls into an impossible local region until the threshold is reached.\\ 
\textbf{ReVoLT w/o distance penalty.} In Fig. \ref{Ablation} (c), using only priors and bonuses can also complete tasks, but their paths are longer due to the fluctuating thoughts while making decisions.\\
\textbf{ReVoLT with continuous mode.} The left figure of Fig. \ref{Ablation} (d) is the same as the one in Fig. \ref{Results}. However, when searching for the second target in this house, once the robot associates current observations with the memory, it can find the target with a higher success rate. However, this also makes the robot more focused on exploitation and reduces exploration, which may cause it to ignore closer targets and lead to a lower SPL.

To sum up, relationship priors are essential for robots to understand the environment semantics, and it is also the major factor affecting the SR. The UCT bonus and distance penalty contribute to the improvement of SPL. \textit{ReVoLT-c} maintains a long-term scene memory and can get outstanding performance.

\section{Conclusion}
We propose \textit{ReVoLT}, a hierarchical reasoning target-driven navigation framework that combines combinatorial graph relation extraction and online UCT decision operating with a multi-layer topological graph. \textit{ReVoLT} shows better performance on exploiting the prior relationships, and its bandit reasoning is more reasonable and efficient. To bridge the gap between existing point-goal controllers and our reasoner, we adopt the Voronoi local graph for the semantic-spatial transition. However, some significant challenges remain in this field. Our future direction lies in using representation learning techniques to introduce richer object information like shape, color, and size, using scene graph detection to introduce richer semantic relation information like furniture layout, and achieving more abundant tasks like object instance navigation.

\bibliographystyle{ieeetr}
\bibliography{ref}

\newpage
\onecolumn
\end{document}